\documentclass[10pt,twocolumn,letterpaper]{article}

\usepackage{cvpr}              

\usepackage[dvipsnames]{xcolor}

\definecolor{cvprblue}{rgb}{0.21,0.49,0.74}
\usepackage[pagebackref,breaklinks,colorlinks,citecolor=cvprblue]{hyperref}
\usepackage{bm}
\usepackage{multirow}
\usepackage{float}
\usepackage{amsmath}
\usepackage{flushend}
\usepackage{booktabs}
\usepackage{colortbl}
\usepackage[accsupp]{axessibility}
\def\paperID{1435} 
\def\confName{CVPR}
\def\confYear{2024}

\title{LEAP-VO: Long-term Effective Any Point Tracking for Visual Odometry}

\author{
Weirong Chen\textsuperscript{1,2}\thanks{\footnotesize Work done at Microsoft Mixed Reality \& AI Lab Zurich for a Master’s degree.} \quad 
Le Chen\textsuperscript{3} \quad 
Rui Wang\textsuperscript{4} \quad 
Marc Pollefeys\textsuperscript{4} \\[3pt]
{\textsuperscript{1}TU Munich}
\,
{\textsuperscript{2}Munich Center for Machine Learning}
\,
{\textsuperscript{3}MPI for Intelligent Systems}
\,
{\textsuperscript{4}Microsoft}
\\[3pt]
}

\begin{document}
\maketitle
\begin{abstract}

Visual odometry estimates the motion of a moving camera based on visual input. Existing methods, mostly focusing on two-view point tracking, often ignore the rich temporal context in the image sequence, thereby overlooking the global motion patterns and providing no assessment of the full trajectory reliability. These shortcomings hinder performance in scenarios with occlusion, dynamic objects, and low-texture areas. To address these challenges, we present the Long-term Effective Any Point Tracking (LEAP) module. LEAP innovatively combines visual, inter-track, and temporal cues with mindfully selected anchors for dynamic track estimation. Moreover, LEAP's temporal probabilistic formulation integrates distribution updates into a learnable iterative refinement module to reason about point-wise uncertainty. Based on these traits, we develop LEAP-VO, a robust visual odometry system adept at handling occlusions and dynamic scenes. Our mindful integration showcases a novel practice by employing long-term point tracking as the front-end. Extensive experiments demonstrate that the proposed pipeline significantly outperforms existing baselines across various visual odometry benchmarks.

%


\end{abstract}    
\section{Introduction}
\label{sec:intro}


%
Visual odometry (VO) calculates the camera's position and movement from an image sequence. It is widely used in various fields including robotics~\cite{yousif2015overview}, mixed reality~\cite{mossel2016streaming}, and autonomous driving~\cite{geiger2013vision}. 
The performance of VO highly depends on the accuracy of its point tracking front-end, which recovers relative camera motion between consecutive frames using projective geometry. It further requires the point correspondences to be static, as dynamic points can lead to inaccurate motion estimation and pose recovery.

\begin{figure}
     \centering
     \setlength{\abovecaptionskip}{0.1cm}
     \setlength{\belowcaptionskip}{-0.2cm}
     \includegraphics[width=0.45\textwidth]{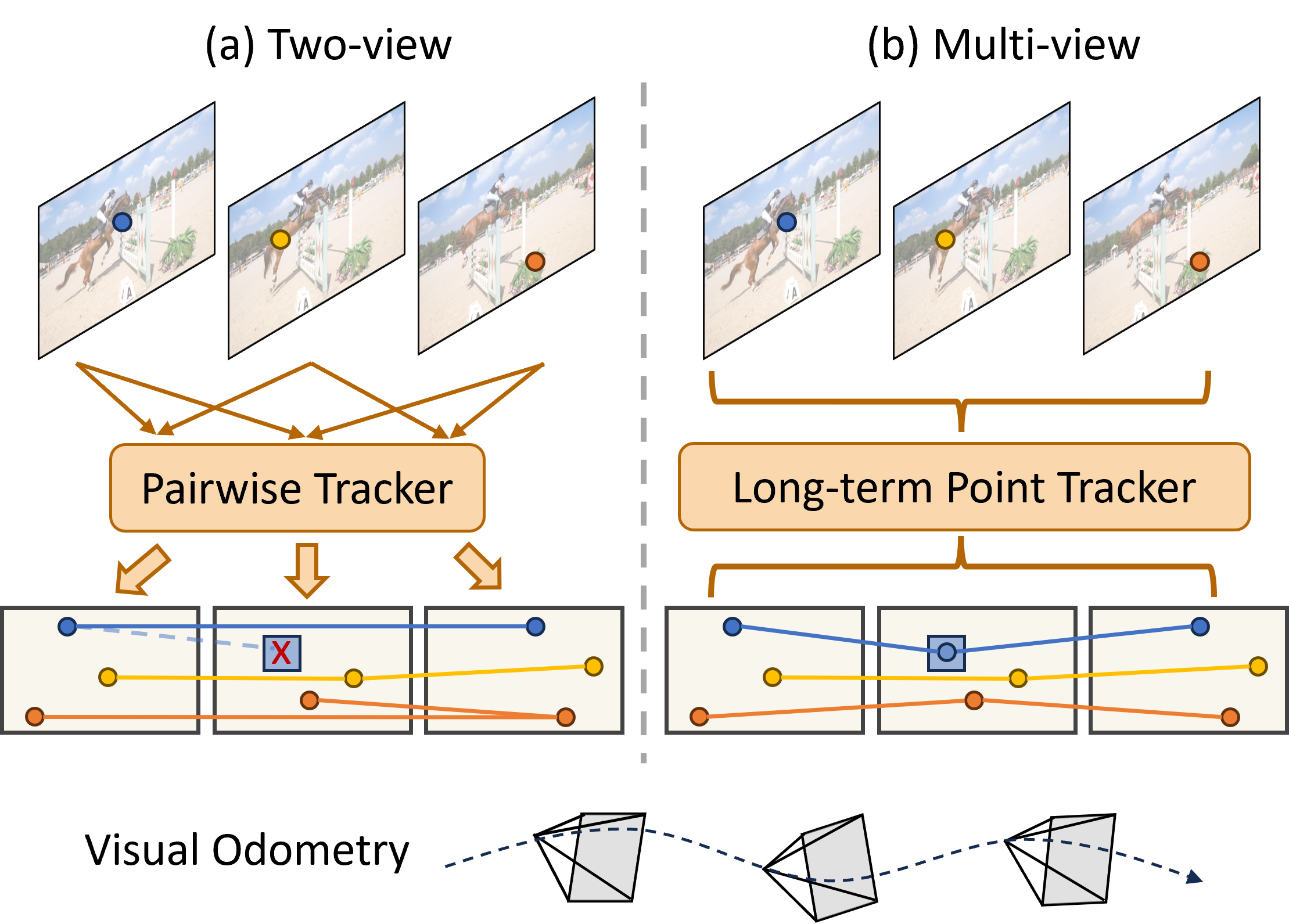}
     \caption{
     \small
     \textbf{Comparison between two-view approach and multi-view approach for visual odometry}. 
     (a) In the two-view approach, correspondences are derived for every image pair and concatenated. Managing occlusion becomes challenging without the temporal context.
    (b) In the multi-view approach, long-term point trajectories can be obtained all at once, enabling the detection of occlusion and tracking points under partial occlusions. 
     }
     \label{fig:teaser}
    \vspace{-0.45cm}
\end{figure}

Despite significant advances in feature tracking and optical flow estimation, most existing VO methods mainly focus on the two-view case, \ie, computing feature correspondences between a pair of images. However, these methods are limited as they rely solely on pairwise relative motion, thus neglecting the valuable temporal information in image sequences. Consequently, they often struggle with capturing dynamic motion patterns in tracks. Another challenge arises from the need to handle occlusions. In two-view scenarios, occlusion is not explicitly modeled, leading to misidentification of occluded points as incorrect matches. 
To address these challenges effectively, we propose to introduce a long-term point tracker, which can recover the trajectories of specific points across the image sequence given a sparse set of these query points. Its capability of leveraging temporal context and reliably tracking queries over multiple frames even under occlusion, makes it highly suitable for VO applications (refer to ~\Cref{fig:teaser}).
Recently, Persistent Independent Particles (PIPs)~\cite{harley2022particle} proposes a novel pipeline that brings the idea of iterative refinement into the long-term point tracking task, with the local correlation evidence to maintain feasible computation. 
%
%
%
While it is proficient in leveraging the temporal context present in the image sequence, it ignores the inter-track information in the image sequence and processes each query point independently. Consequently, PIPs struggles to capture the global motion patterns from the image sequence. 
Additionally, it lacks a reliability metric for its trajectory estimates, hindering its application in quality-sensitive tasks like video segmentation and visual odometry.
%
%
To mitigate the aforementioned issues, we propose a pipeline that explores dynamic track estimation and temporal probabilistic formulation for long-term point tracking. 

In this paper, we present Long-term Effective Point Tracking (LEAP), aiming to estimate the reliability of predicted correspondences and track moving object trajectories in dynamic scenes.
%
To achieve this, we harness the global motion patterns—often overlooked in existing methods—through an anchor-based dynamic track estimation module, leveraging visual, temporal, and inter-track information to differentiate between the static and dynamic elements. We also incorporate a temporal probabilistic approach into our LEAP pipeline to handle uncertainties and refine the distribution of point correspondences iteratively, showcasing superior performance in complex real-world situations.
More importantly, we pioneer the usage of continuous motion estimation from long-term point tracking to construct a robust visual odometry system. Our LEAP-VO distinguishes itself through its thoughtful integration of visual features, track distribution, and global motion patterns, significantly enhancing both the performance and robustness of visual odometry in dynamic environments.
%
%
%

\section{Related Work}
\label{sec:related_work}

\paragraph{Long-term Point Tracking.}

In the realm of long-term point tracking or Tracking Any Point (TAP), various methods have been introduced recently. 
PIPs~\cite{harley2022particle} reframes pixel tracking as a long-term motion estimation with local information and improves tracking accuracy by associating each pixel with a trajectory that spans multiple image frames. 
%
%
The subsequent works enhance performance by integrating spatial context features~\cite{bian2023contexttap}, tracking multiple queries concurrently~\cite{karaev2023cotracker}, and extending temporal fields and templates~\cite{zheng2023pointodyssey}.
TAP-Net~\cite{doersch2022tap} adopts a different method by locating candidate correspondences through a global cost volume, improved by TAPIR's trajectory refinement~\cite{doersch2023tapir} and BootsTAP's semi-supervised training~\cite{doersch2024bootstap}.
Other approaches explore self-supervision targets to learn visual correspondence via powerful feature representation~\cite{jabri2020space, caron2021emerging, tumanyan2024dino}. 
%
%
Recently, OmniMotion~\cite{wang2023omnimotion} provides a pure test-time optimization method that infers the point trajectory and occlusion directly from videos with optical flow predictions.
%
However,
its slow inference speed makes it unsuitable for real-time applications such as visual odometry. 

\paragraph{Monocular Visual Odometry.}
%
%
%
Visual Odometry (VO) can be categorized into indirect (feature-based) and direct methods. Indirect methods, such as MonoSLAM~\cite{davison2007monoslam} and ORB-SLAM~\cite{mur2015orb}, mostly rely on feature extraction and matching to estimate the camera pose. 
%
%
Direct methods, conversely, skip feature extraction and matching and utilize intensity to optimize camera pose and 3D scene point positions based on photometric errors~\cite{engel2014lsd, engel2017direct, wang2017stereo}, showing better performance on low-texture regions where keypoint tracking can easily fail. However, they are more vulnerable to illumination changes and issues with rolling shutter cameras.
%
Advancements in deep learning have spawned end-to-end monocular VO methods in  supervised~\cite{wang2017deepvo,bloesch2018codeslam, teed2019deepv2d, yang2020d3vo, wang2021tartanvo, teed2022deep, shen2023dytanvo, ye2023pvo} and unsupervised~\cite{zhou2017unsupervised, yin2018geonet, ranjan2019competitive, li2020self} settings.
%
%
DeepVO~\cite{wang2017deepvo} utilizes Recurrent Neural Networks to model sequential data, while SfMLearner~\cite{zhou2017unsupervised} develops an unsupervised framework to learn the depth and camera motion from unlabeled monocular videos. TartanVO~\cite{wang2021tartanvo} proposes a VO model that generalizes to multiple datasets and real-world scenarios. DROID-SLAM~\cite{teed2021droid} follows learning-based optimization from RAFT~\cite{teed2020raft} and proposes an end-to-end system, integrating flow, confidence, and geometric optimization via iterative GRU updates. The followed-up work DPVO~\cite{teed2022deep} further cuts computational complexity by replacing dense feature tracking with sparse patch tracking. 
%
\paragraph{Dynamic Track Estimation.}
In addition to enhancing accuracy, numerous studies have focused on the detection of dynamic objects within videos.  Methods based on segmentation, such as DynaSLAM~\cite{bescos2018dynaslam} and RCVD~\cite{kopf2021robust}, employ semantic cues from pre-trained models to exclude dynamic regions. However, these segmentation-based methods are constrained to predefined object categories and struggle to capture real-world motion (for example, differentiating between a stationary and a moving car). Alternatively, DytanVO~\cite{shen2023dytanvo} and ParticleSfM~\cite{zhao2022particlesfm} adopt a trajectory-based motion detection approach. This method addresses dynamic scenes using estimated point track information, thereby generating reliable camera trajectories in complex motion scenarios. Nonetheless, they depend on dense optical flow estimation to capture global motion patterns.  Furthermore, dynamic tracks can be detected using additional depth information for a 3D consistency check~\cite{chen2019self, zhang2022structure}, or by optimization based on photo consistency~\cite{tschernezki2021neuraldiff, wu2022d, yang2023emernerf}.

%
%
%
%
%
%
%

\section{Method}
\label{sec:method}

Our goal is to tackle the aforementioned challenges in VO systems, particularly in handling dynamic environments, temporal occlusions, and low-texture regions. To this end, we introduce our LEAP module, which incorporates dynamic track estimation and temporal probabilistic formulation. Furthermore, we illustrate how this module serves as the cornerstone for constructing our advanced VO system.


\subsection{Preliminary: Tracking Any Point}
\label{sec:method-preliminary}

Tracking arbitrary pixels in an image sequence is traditionally akin to estimating optical flow, where each pixel is mapped to a corresponding location in a subsequent frame. 
However, two-view approaches often overlook the rich temporal context within the image sequence, leading to suboptimal outcomes. Recent advancements address this by revisiting the particle video concept and improving long-term point tracking through learning-based techniques~\cite{harley2022particle, doersch2023tapir, karaev2023cotracker}. 
%
These TAP methods can track points over multiple frames and detect occlusions, significantly boosting tracking robustness under challenging scenarios. \\

\noindent
\textbf{Formulation}.
Given a sequence of $S$ consecutive RGB images $\mathbf{I}=[\mathbf{I}_1,...,\mathbf{I}_S], \mathbf{I}_s \in \mathbb{R}^{3\times H\times W}$ as input, the goal of TAP is to track a set of query points 
across these frames.
%
For a specific query point with the 2D pixel coordinate
$\mathbf{x}_q \in \mathbb{R}^2$ in frame $s_q$, TAP predicts its trajectory across all $S$ images as $\mathbf{X} = [\mathbf{x}_1, ..., \mathbf{x}_{S}], \mathbf{x}_s \in \mathbb{R}^2$, given $\mathbf{x}_{s_q} = \mathbf{x}_q$. 
Additionally,
TAP estimates the visibility of each point throughout the sequence as $\mathbf{V} = [v_1,..., v_{S}]$, where $v_s \in \{0,1\}$ is a binary label of point visibility indicating whether a point in frame $s$ is visible or occluded. The TAP formulation for a single query can be described as 
\begin{equation}
    (\mathbf{X}, \mathbf{V}) = \text{TAP}(\mathbf{I}, \mathbf{x}_q, s_q).
\end{equation}

PIPs~\cite{harley2022particle}, a notable learning-based TAP method, offers a promising solution featuring local cost volumes and iterative refinement. Initially, each image $\mathbf{I}_s \in \mathbf{I}$ is processed through a CNN feature extractor $\mathcal{F}$ to extract the feature map $\mathbf{Y_s} = \mathcal{F}(\mathbf{I}_s)$. Point features $\mathbf{f}_q$ are computed by bilinear sampling at the query positions $\mathbf{x_s}$ on $\mathbf{Y}_s$. 
These point features from each image are then concatenated to form a point feature tensor
$\mathbf{F} = [\mathbf{f}_1, \ldots, \mathbf{f}_S]$. PIPs approaches the TAP problem by iteratively updating the state variables $(\mathbf{X}, \mathbf{F})$, with the initial states obtained by duplicating the query positions and features. 
During the $k$-th iterative refinement, PIPs computes a multi-scale local cost volume $\mathbf{C}[\mathbf{X}^k]$ centered around the current estimated point position $\mathbf{X}^k$. Utilizing this local evidence, PIPs predicts updates for the state variables via 
\begin{equation}
\begin{split}
    (\Delta \mathbf{X}, \Delta \mathbf{F}) &= \text{Refiner}(\mathbf{F}^k, \text{pos}(\mathbf{X}^k - \mathbf{x_q}), \mathbf{C}^k [\mathbf{X}^k]),\\
    \mathbf{X}^{k+1} &\leftarrow \mathbf{X}^k + \Delta \mathbf{X}, \quad
    \mathbf{F}^{k+1} \leftarrow \mathbf{F}^k + \Delta \mathbf{F},
    \end{split}
\end{equation}
where $\text{pos}(\cdot)$ refers to the positional embedding. The final point feature $\mathbf{F}^K$ is used to predict point visibility with a simple linear projection layer $\mathcal{G}_v$, yielding $\mathbf{V} = \mathcal{G}_v(\mathbf{F}^K)$. 
Recent advances extend PIPs by replacing the MLP-based refiner with an attention-based refiner and adding spatial feature aggregation to track multiple queries together~\cite{karaev2023cotracker}.

\begin{figure}[]
 \centering
     \setlength{\abovecaptionskip}{0.1cm}
     \setlength{\belowcaptionskip}{-0.4cm}
      \begin{subfigure}[b]{0.47\textwidth}
         \centering
         \includegraphics[width=\textwidth]{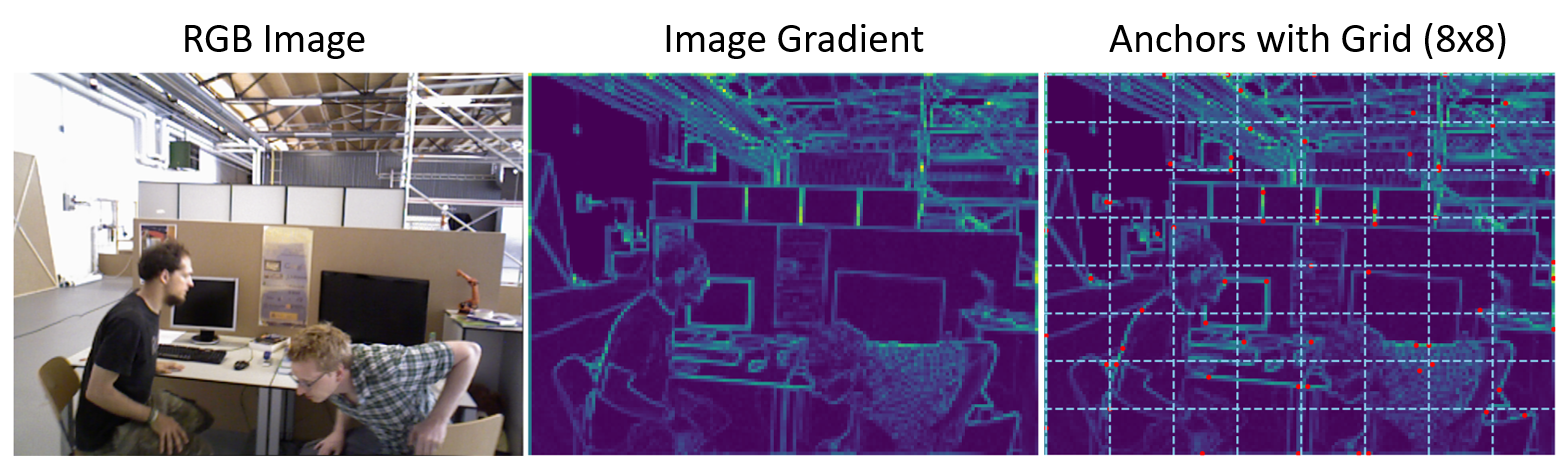}
     \end{subfigure}
    \caption{
   \textbf{Distributed image gradient-based sampling    with $k=8, N_a=64$.}
     After computing the image gradient and pooling, we split the gradient map into 8×8 grids and select the point with the maximum gradient in each grid.
    }
     \label{fig:max_grid_sampling}
    \label{fig:mixer-all}
    \vspace{-5px}
\end{figure}

\subsection{Anchor-based Dynamic Track Estimation}
\label{sec:method_dynamic_track}
The established TAP model enables us to extract long-term point trajectories and thereby recover camera motion from an image sequence.  
However, dynamic environments pose extra challenges for VO systems that rely on static 3D point correspondences linked to unchanging scene locations.
Therefore, integrating dynamic track estimation into the VO system is crucial for real-world applications.

We propose an innovative method for dynamic track estimation that leverages visual, temporal, and inter-track information. While common techniques like monocular segmentation~\cite{kopf2021robust} 
excel at identifying moving objects using visual cues but can erroneously label static objects, like parked cars, as dynamic. 
Trajectory-based methods~\cite{zhao2022particlesfm}, utilizing optical flows from two or multiple frames, can differentiate static and moving objects apart but tend to ignore visual information and can be inefficient due to dense trajectories computation. Merging these two distinct methods is nontrivial; however, the formulation of the TAP pipeline, enhanced with inter-track attention, offers a viable solution.

Our method smartly integrates visual appearance, long-term motion, and inter-track cues, combining the strengths of segmentation and trajectory-based techniques. Utilizing TAP, we extract point features $\mathbf{F}$ and long-term trajectories $\mathbf{X}$ from the video input. These features capture visual information and the iterative refinement module exchanges information across tracks over time for accurate labeling. 
However, such a method depends on the query distribution and quantities. 
With only one query during inference time, it cannot utilize inter-track information.
%
%
Likewise, if the queries focus on a single object, the system may overlook the motions of other objects due to similar motion patterns. \\

\noindent
\textbf{Anchor-based Inter-track Attention}.
To better capture global motion patterns, we introduce additional anchors alongside the given queries, denoted as $\mathcal{A}, |\mathcal{A}|=N_a$. Ideally, the anchor set should exhibit two characteristics: (1) the anchors should be easy to track, and (2) they should be well-distributed to encapsulate global motion patterns. Given the image $\mathbf{I}_{s_q}$ from which the queries ${\mathbf{x}_q }$ are extracted (we maintain the same notations for both single and multiple queries for simplicity), we first employ the Sobel Kernel to derive the image gradient $(\mathbf{G}_x, \mathbf{G}_y)$ for image $\mathbf{I}_{s_q}$.
We then apply an average pooling layer on gradient maps for smoothing and downscaling feature maps. Subsequently, we divide the gradient map into $k\times k$ sub-regions and select top $\frac{N_a}{k^2}$ pixels with the highest image gradient magnitude in each local grid. 
The process is demonstrated in \Cref{fig:max_grid_sampling}. 

To predict the dynamic label of a trajectory, we track queries with additional anchors and concatenate the original query features $\mathbf{X}$ and query trajectories $\mathbf{F}$ with anchor features $\mathbf{X}_A$ and anchor trajectories $\mathbf{F}_A$. We apply a shallow MLP layer $\mathcal{G}_d$ with average pooling over the temporal dimension. The dynamic track label $\mathbf{m}_{d}$ is estimated as  
\begin{equation}
    \mathbf{m}_{d} = \text{avgpool}(\mathcal{G}_d([\mathbf{X}^K; \mathbf{X}^K_A], [\mathbf{F}^K; \mathbf{F}^K_A])).
\end{equation}

Since newly added anchors often lack ground truth information, we employ a semi-supervised training scheme that tracks queries and anchors together to leverage inter-track attention but only calculates losses based on query predictions. This approach prevents the network from developing bias towards specific query sampling methods while learning dynamic track labels effectively.

\subsection{Temporal Probability Modeling}
\label{sec:method_probability}
Supervised learning applied to large-scale datasets has yielded notable advancements in learning-based point tracking methods. Nonetheless, 
%
%
trajectory estimation errors remain a concern, especially under challenging conditions like occlusions, motion blur, or textureless regions. 
%
%
Recognizing the varying reliability of trajectories, 
integrating an uncertainty estimation mechanism into the correspondence estimation process is beneficial~\cite{chen2023uncertainty}. 
This is particularly crucial for quality-sensitive tasks such as visual odometry.

To evaluate the reliability of point correspondences, we incorporate a probabilistic formulation into the TAP pipeline. Given an image sequence $\mathbf{I}$ and a single query point $\mathbf{x}_q$, our objective is to compute the conditional probability density of point trajectory as $p(\mathbf{X}|\mathbf{I}, \mathbf{x}_q)$. Recognizing the strong correlation between points within the same track, treating each point in the track as independent becomes suboptimal. Therefore, we propose to model the distribution of point trajectories by applying two multivariate distributions, respectively for the 2D coordinates. Let $\mathbf{X} = [\mathbf{a};\mathbf{b}]$,  where $\mathbf{a} \in \mathbb{R}^S$ and $\mathbf{b} \in \mathbb{R}^S$ represent the X  and Y coordinate of all points in $\mathbf{X}$, respectively.  Assuming independence between these two coordinates, the joint probability distribution can be expressed as $p(\mathbf{X}|\mathbf{I}, \mathbf{x}_q) = p(\mathbf{a}|\mathbf{I}, \mathbf{x}_q) \cdot p(\mathbf{b}|\mathbf{I}, \mathbf{x}_q)$. 
We adopt the multivariate Cauchy distribution, which is associated with its heavy tails distribution and is more stable for optimization. 
The probability density function (PDF) of a single coordinate is given by
\begin{equation}
p(\mathbf{a}|\mathbf{I}, \mathbf{x}_q)\!=\!\frac{\Gamma(\frac{1+S}{2})}{\Gamma(\frac{1}{2})\pi^\frac{S}{2} |\mathbf{\Sigma}_a|^{\frac{1}{2}}[1\!+\!(\mathbf{\mathbf{a}}\!-\!\bm{\mu}_a)^T \mathbf{\Sigma}_a^{-1}(\mathbf{\mathbf{a}}\!-\!\bm{\mu}_a)]^{\frac{1+S}{2}}}
\end{equation}
and a similar expression is applicable for $p(\mathbf{b}|\mathbf{I}, \mathbf{x}_q)$. 
$\Gamma$ is the Gamma function.
The parameters  $(\bm{\mu}_a, \mathbf{\Sigma}_a, \bm{\mu}_b, \mathbf{\Sigma}_b)$ 
represent the location and scale matrices for the respective coordinates.
%
%
During inference, the uncertainty associated with each point is quantified by the sum of diagonal scale estimates at position $(s,s)$ as
$\phi(\mathbf{x}_s) = \mathbf{\Sigma}_{a}[s,s] + \mathbf{\Sigma}_{b}[s,s]$.\\

\noindent
\textbf{Kernel-based Estimation}.
Rather than just predicting the positions of points within a trajectory, our model focuses on recovering the distribution of point trajectories by estimating its parameters.
The location parameter predictions can be associated with the previous point trajectory as $[\bm{\mu}_a; \bm{\mu}_b] = \mathbf{X}$. However, directly deriving a symmetric and positive definite scale matrix from model outputs is challenging. To address this, we employ a kernel-based approach with the linear kernel $K(\mathbf{x},\mathbf{y}) = \mathbf{x}^T \mathbf{y}$. During each iteration, two scale matrices $(\mathbf{\Sigma}_a, \mathbf{\Sigma}_b)$ are constructed by first applying two linear projection layers to the point features, represented as  $\mathbf{F}_a^k = \mathcal{G}_a(\mathbf{F}^k)$ and $\mathbf{F}_b^k = \mathcal{G}_b(\mathbf{F}^k)$. We then compute the scale matrices as $\mathbf{\Sigma}_a = K(\mathbf{F}_a, \mathbf{F}_a) + \sigma \mathbf{I}$, and $\mathbf{\Sigma}_b = K(\mathbf{F}_b, \mathbf{F}_b)+ \sigma \mathbf{I}$, where $\sigma$ is a small positive value and $\mathbf{I}$ is the identity matrix. The model parameters are refined through Maximum Likelihood Estimation (MLE) using a negative log-likelihood (NLL) loss function. \\

\begin{figure}[]
     \centering
     \setlength{\abovecaptionskip}{0.01cm}
     \setlength{\belowcaptionskip}{-0.45cm}
   \begin{subfigure}[]{0.47\textwidth}
         \centering
         \includegraphics[width=\textwidth]{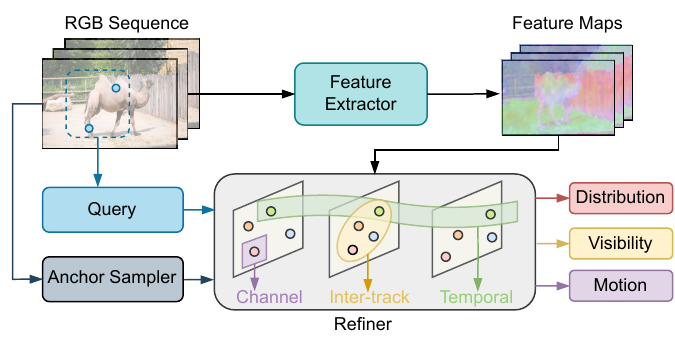}
     \end{subfigure}
     
    \caption{
    \textbf{LEAP Front-end}. Once image feature maps are obtained, selected anchors aid in tracking. The queries and anchors are processed by a refiner to iteratively update states. The model outputs trajectory distribution, visibility, and dynamic track label.
    }
    \label{fig:leap_frontend}
\end{figure}

\begin{figure*}[h]
    \centering
     \setlength{\abovecaptionskip}{0.1cm}
     \setlength{\belowcaptionskip}{-0.5cm}
    \includegraphics[width=0.91\textwidth]{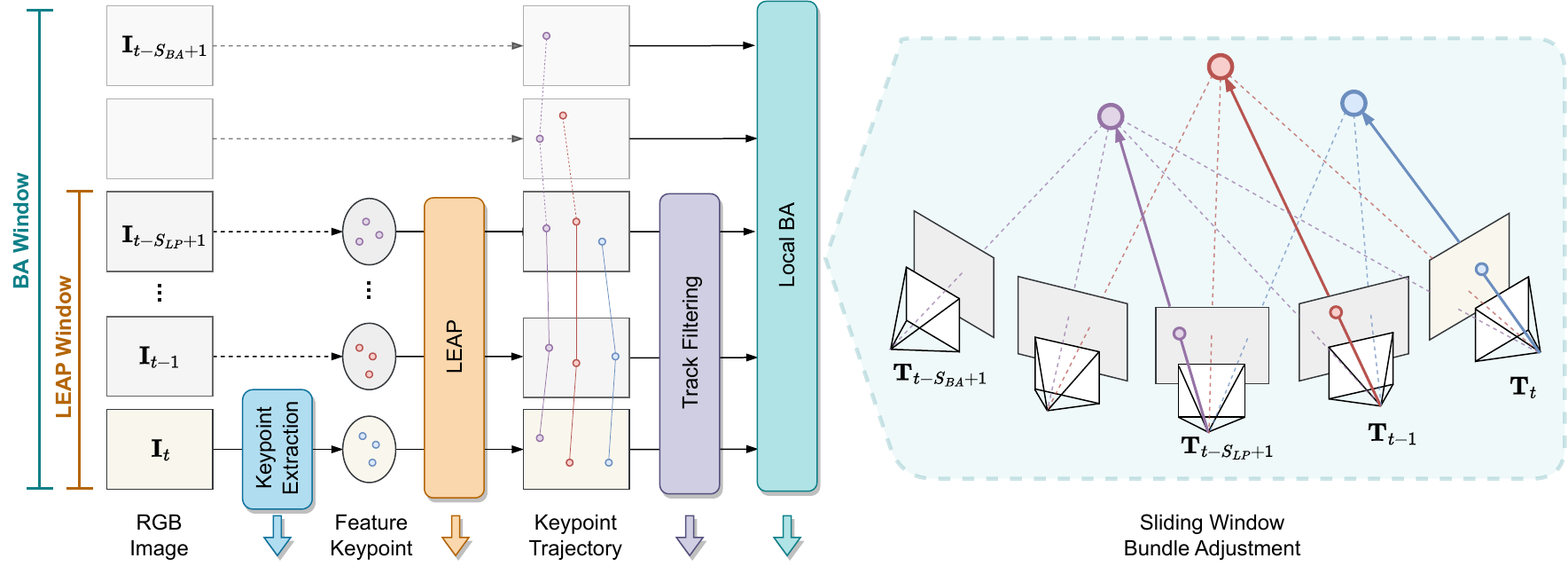}
    \caption{
    \small
    \textbf{LEAP-VO pipeline.} Given a new image $\mathbf{I}_t$ received at time $t$, the keypoint extractor extracts new keypoints from $\mathbf{I}_t$. Then, all the keypoints from the latest $S_{LP}$-frame $\mathbf{I}_{t-S_{LP}+1:t}$ are tracked across all other frames within the current LEAP window, followed by a track filtering step to remove outliers. Finally, the local BA module is used on the current BA window to update the camera poses and 3D positions of the extracted keypoints. The colored arrows denote the moving direction of each module when a new image is received. 
    }
    \label{fig:leap-vo}
\end{figure*}

\noindent
\textbf{Loss Functions}.
We supervise the point trajectory mainly with the NLL loss, which is based on the predicted distribution parameters and the ground-truth point trajectory. 
The main point trajectory loss is given as
\begin{align}
    \mathcal{L}_{main} = \sum_k^K \gamma^{K-k} \mathcal{L}_{NLL}(\mathbf{X}^k, \mathbf{X^*}, \mathbf{\Sigma}_a^k, \mathbf{\Sigma}_b^k),
\end{align}
where $K$ is the number of iteration and $\gamma=0.8$. 

For the visibility and dynamic track label supervision, we use the cross entropy loss with the estimates $\mathbf{V}, \mathbf{m}_{d}$ and ground truth $\mathbf{V}^*, \mathbf{m}_{d}^*$ as 
\begin{align}
    &\mathcal{L}_{vis} = (1 - \mathbf{V}^*) \log(1 - \mathbf{V}) + \mathbf{V}^* \log \mathbf{V}. \\
    &\mathcal{L}_{dyn} = (1 - \mathbf{m}_{d}^*) \log(1 - \mathbf{m}_{d}) + \mathbf{m}_{d}^* \log \mathbf{m}_{d}.
\end{align}
The total loss is a weighted sum of three losses:
\begin{align}
    \mathcal{L}_{total} = w_1 \mathcal{L}_{main} + w_2 \mathcal{L}_{vis} + w_3 \mathcal{L}_{dyn}.
\end{align}

\subsection{LEAP-VO}
\label{sec:method_leap_vo}
Leveraging the anchor-based dynamic track estimations and temporal probabilistic formulation, we develop our Long-term Effective Point Tracking (LEAP) pipeline, boosting the accuracy of point tracking and simultaneously capturing global motion patterns and uncertainties of the tracks (refer to \Cref{fig:leap_frontend}). In this section, we demonstrate how we can build visual odometry for dynamic scenes based on the long-term tracking capabilities of LEAP. Our LEAP-VO pipeline is illustrated in \Cref{fig:leap-vo}.\\

\noindent
\textbf{Keypoint Extraction and Tracking}.
In our VO system, the LEAP model serves as the front-end. When a new image frame $\mathbf{I}_t$ is captured at time $t$,  we sample $N$ new keypoints from this image to initiate point tracking. We consistently employ the distributed image gradient-based sampling method (refer to \Cref{sec:method_dynamic_track}) for keypoint selection due to its favorable properties in feature distinctiveness and distribution. This approach also obviates the need for additional anchors during tracking.  The LEAP model then tracks these keypoints across the last $S_{LP}$ frames, performing bidirectional tracking (both forward and backward in time). 
Notably, $S_{LP}$ can exceed the model window $S$, thus enabling more expansive tracking capabilities through sequential prediction~\cite{harley2022particle, karaev2023cotracker}. \\

\noindent
\textbf{Track Filtering}.
Theoretically, precise camera pose estimation between two images can be attained with just a few accurate correspondences under certain conditions. However, the quality of correspondences obtained from feature matching or optical flow can vary. Traditional VO pipelines often use RANSAC~\cite{fischler1981ransac} to filter out incorrect matches, but their efficiency decreases with more keypoints and views. Our approach, in contrast, bypasses the sampling-based method and uses trajectory quality assessments from LEAP for more efficient and informative track filtering. 

\textit{(i) Visibility and Dynamic Track Label:}
The LEAP model has the capability to estimate not only the 2D point positions within a trajectory but also the visibility of each point and the dynamic track label of the entire trajectory. Thresholds  $\gamma_{v}$ for visibility and $\gamma_{d}$ for dynamic track labels are set to ensure that only visible and static points are utilized in the bundle adjustment process.



\textit{(ii) Track Uncertainty}:
In \Cref{sec:method_probability}, we introduce probabilistic modeling to LEAP, allowing us to gauge confidence via distribution measurements. We select only high-confidence points, which correspond to low uncertainty measurements. Denoting our model's uncertainty estimation as $\mathbf{\Phi}$, we retain points that exhibit high confidence using the criterion $\mathbf{\Phi} \leq Q(\gamma_{u})$, where $\gamma_u$ is the uncertainty quantile and $Q:[0,1]\rightarrow \mathbb{R}$ is the quantile function. 

\textit{(iii) Track Length}: 
After filtering, we evaluate and remove tracks with insufficient observations, as bundle adjustment is more reliable with increased observations and wider baselines. Tracks with fewer than $\gamma_{track}$ valid points are excluded from optimization cost computation due to their potential unreliability.\\
\noindent
\textbf{Sliding-window Optimization}.
With the point validation mask established, we proceed to define the optimization cost function for our sliding window bundle adjustment. The window size for local bundle adjustment, denoted as $S_{BA}$, may differ from the point tracking window size  $S_{LP}$. Our geometric bundle adjustment aims to optimize camera poses $\mathbf{T}$ and 3D scene point positions $\mathbf{Q}$ by aligning the induced point trajectory from the projective relationship, with the estimated point trajectory from LEAP. 
 We parameterize a 3D scene point $\mathbf{Q}_i$ by its 2D keypoint location $\mathbf{x}_i$ in image $\mathbf{I}_i$ and depth $d_i$. Let $\text{LEAP}_{i\rightarrow j}$ denote the mapping of tracking keypoint from $\mathbf{I}_i$ to $\mathbf{I}_j$ using LEAP model. The reprojection cost is formulated as:
\begin{align}
\label{eq:ba}
    \sum_i \sum_{j\in |i-j| \leq S_{BA}} \sum_{n} w_{i \rightarrow j,n} \|
    \mathcal{P}(\mathbf{T}_i, \mathbf{T}_j, \mathbf{K}, d_{i,n}) - \nonumber \\ 
    \text{LEAP}_{i \rightarrow j} (\mathbf{x}_{i,n})
    \|_{\rho},
\end{align}
where $\| \cdot \|_{\rho}$ is the distance metric, and the weight $w_{i \rightarrow j,n}$ is derived from the track filtering results. 
We use the Gauss-Newton method~\cite{triggs1999bundle} to optimize \cref{eq:ba} for $K_{BA}$ iterations. 
At each iteration, we compute the camera poses update $\Delta \xi^{(k)} \in \mathfrak{se}(3)~\textrm{(lie-algebra corresponding to} ~\mathbf{T})$, and the depth update $\Delta \mathbf{D}^{(k)}$ for each point. The optimization can be solved efficiently with the Schur decomposition. We use the robust Huber loss function for the distance metric. 
\section{Experiment}
\label{sec:experiment}

\subsection{Datasets and Metrics}
\textbf{Replica}. 
Replica~\cite{straub2019replica} provides synthetic indoor environments for simulating the image-capturing process with a moving camera. We use the camera trajectories provided by~\cite{zhi2021place} for the localization task, resulting in 16 camera trajectories derived from 8 static scenes. 
Each sequence comprises 900 frames with rendered RGB images. We use ``Sequence 1" for the visual odometry evaluation. \\[-4pt]

\noindent
\textbf{MPI Sintel}.
The MPI Sintel~\cite{Butler2012sintel} dataset contains dynamic image sequences from open-source 3D animated short films. These sequences consist of large and complex object motion, along with other imaging effects such as motion blur and defocus blur. 
Following prior works~\cite{kopf2021robust, zhao2022particlesfm}, we assess dynamic VO performance on the MPI Sintel dataset, with each sequence containing 20 to 50 image frames.  \\[-4pt]

\noindent
\textbf{TartanAir Shibuya}. 
The TartanAir Shibuya dataset~\cite{qiu2022airdos} features dynamic scenes from two scenarios: ``Standing Humans" and ``Road Crossing". 
The majority of the humans remain stationary in "Standing Humans" sequences, while  "Road Crossing" sequences provide a more challenging setting where multiple humans are moving in different directions. 
Each sequence in the dataset includes 100 frames with more than 30 tracked moving humans.\\[-5pt]

\noindent
\textbf{Metrics}. 
For visual odometry, we compare the Absolute Translation Error (ATE), Relative Translation Error (RPE trans), and Relative Rotation Error (RPE rot) following ~\cite{zhao2022particlesfm, zhang2022structure, teed2022deep}. ATE measures how much the estimated trajectory deviated from the ground truth trajectory, which is calculated as the root mean square error (RMSE) overall all pose transformations after scaling and alignment. ``RPE trans'' computes translation errors made over a certain distance (meter), and ``RPE rot'' computes the estimated rotation over a certain distance (degree). Both ``RPE trans'' and ``RPE rot'' are calculated on all poses and then averaged.

\begin{table}[]
\centering
\footnotesize
 \setlength{\abovecaptionskip}{0.1cm}
 \setlength{\belowcaptionskip}{-0.1cm}
\setlength\tabcolsep{8pt}
\begin{tabular}{@{}lccc@{}}
\toprule
\multicolumn{1}{c}{\multirow{2}{*}{Method}} & \multicolumn{3}{c}{Replica}                   \\ \cmidrule(l){2-4} 
\multicolumn{1}{c}{}                        & ATE (m)        & RPE trans (m)  & RPE rot (deg)  \\ \midrule
ORB-SLAM2~\cite{murorbslam2}          &   0.086         &  0.030          &  0.650             \\
DynaSLAM~\cite{bescos2018dynaslam}                              &   0.039             &    0.017            &     0.366           \\
DROID-SLAM~\cite{teed2021droid}      &  0.267         &   0.036       &    2.631       \\ \midrule
TartanVO~\cite{wang2021tartanvo}       &  0.406         &  0.036       &  2.063     \\
DytanVO~\cite{shen2023dytanvo}                           &      0.289              &     0.035              &    2.146            \\
DPVO~\cite{teed2022deep}            & 0.257 & 0.036         & 2.635         \\ 
\textbf{LEAP-VO (Ours)}     & \textbf{0.204}  &\textbf{0.030} &  \textbf{1.992}\\ \bottomrule
\end{tabular}
\caption{\textbf{Camera tracking results on Replica~\cite{straub2019replica}.}
The statistics demonstrate that our method achieves better results than other competitive baselines.
}
\label{tab:vo_replica}
\end{table}
\begin{table}[]
\centering
\footnotesize
 \setlength{\abovecaptionskip}{0.1cm}
 \setlength{\belowcaptionskip}{-0.5cm}
\setlength\tabcolsep{8pt}
\begin{tabular}{@{}lccc@{}}
\toprule
\multicolumn{1}{c}{\multirow{2}{*}{Method}} & \multicolumn{3}{c}{MPI Sintel}                   \\ \cmidrule(l){2-4} 
\multicolumn{1}{c}{}                        & ATE (m)        & RPE trans (m)  & RPE rot (deg)  \\ \midrule
ORB-SLAM2~\cite{murorbslam2}          & X              & X              & X              \\
DynaSLAM~\cite{bescos2018dynaslam}    &  X              &   X             &    X            \\
DROID-SLAM~\cite{teed2021droid}      & 0.175          & 0.084          & 1.912          \\ \midrule
TartanVO~\cite{wang2021tartanvo}       & 0.238          & 0.093          & 1.305          \\
DytanVO~\cite{shen2023dytanvo}                           &    0.131                &   0.097                 &   1.538             \\
DPVO~\cite{teed2022deep}            & 0.076 & 0.078         & 1.722         \\ 
\textbf{LEAP-VO (Ours)}     & \textbf{0.037} & \textbf{0.055} & \textbf{1.263} \\ \bottomrule
\end{tabular}
\caption{\textbf{Camera tracking results on MPI Sintel~\cite{Butler2012sintel}.}
The statistics show that our method outperforms other competitive baselines.`X' denotes the tracking failure case. }
\label{tab:vo_sintel}
\end{table}


\begin{table*}[ht]
\centering
\footnotesize
 \setlength{\abovecaptionskip}{0.1cm}
 \setlength{\belowcaptionskip}{-0.15cm}
\setlength\tabcolsep{10pt}
\begin{tabular}{@{}lcccccccc@{}}
\toprule
\multicolumn{1}{c}{\multirow{2}{*}{Method}} & \multicolumn{2}{c}{StandingHuman} & \multicolumn{3}{c}{RoadCrossing (Easy)}             & \multicolumn{2}{c}{RoadCrossing (Hard)} & \multirow{2}{*}{Average} \\ \cmidrule(lr){2-8}
\multicolumn{1}{c}{}                        & 01              & 02              & 03              & 04              & 05              & 06                 & 07                 &                          \\ \midrule
ORB-SLAM2~\cite{murorbslam2} w/ mask                           & 0.0788          & 0.0060          & 0.0657          & 0.0196          & 0.0148          & 1.0984             & 0.8476             & 0.3044                   \\
DynaSLAM~\cite{bescos2018dynaslam}                                    & X               & 0.8836          & 0.3907          & 0.4196          & 0.4925          & 0.6446             & 0.6539             & (0.5808)                 \\
AirDOS~\cite{qiu2022airdos} w/ mask                                 & 0.0606          & 0.0193          & 0.0951         & 0.0331          & 0.0206          & 0.2230             & 0.5625            & 0.1449                   \\
DROID-SLAM~\cite{teed2021droid}                                  & 0.0051          & 0.0073          & 0.0103          & 0.0120          & 0.2278          & 0.0253             & 0.5788             & 0.1238                   \\ \midrule
TartanVO~\cite{wang2021tartanvo}                                    & 0.0600          & 0.1605          & 0.2762          & 0.1814          & 0.2174          & 0.3228             & 0.5009             & 0.2456                   \\
DytanVO~\cite{shen2023dytanvo}                                     & 0.0327          & 0.1017          & 0.0608          & 0.0516          & 0.0755          & \textbf{0.0365}    & 0.0660             & 0.0607                   \\
DPVO~\cite{wang2017deepvo}                                        & 0.0539          & 0.1603          & 0.1788          & 0.2858          & 0.1196          & 0.1119             & 0.1498             & 0.1514                   \\
\textbf{LEAP-VO (Ours)}                                        & \textbf{0.0081} & \textbf{0.0185} & \textbf{0.0399} & \textbf{0.0317} & \textbf{0.0113} & 0.0689         & \textbf{0.0246}    & \textbf{0.0290}          \\ \bottomrule
\end{tabular}
\caption{\textbf{ATE (m) results on TartanAir-Shibuya  Sequences~\cite{qiu2022airdos}.} $(\cdot)$ denotes averaging on valid scenes only. 
We can observe that our method shows better results for ATE on most sequences, outperforming all SLAM and VO baselines on average ATE. 
}
\label{tab:vo_shibuya}
\vspace{-4pt}
\end{table*}
\begin{figure*}[h]
     \centering
     \setlength{\belowcaptionskip}{-0.25cm}
     
    \begin{subfigure}[b]{0.245\textwidth}
         \centering
         \includegraphics[width=\textwidth]{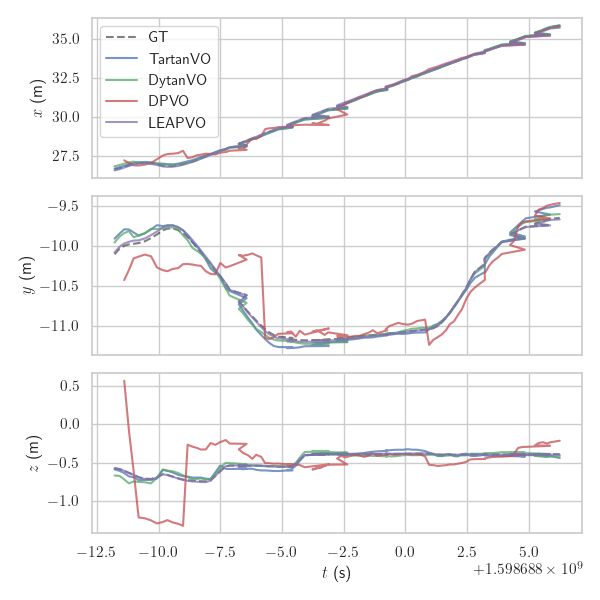}
        \setlength{\abovecaptionskip}{-0.45cm}
         \caption{RoadCrossing 04}
     \end{subfigure}
     \begin{subfigure}[b]{0.245\textwidth}
         \centering
         \includegraphics[width=\textwidth]{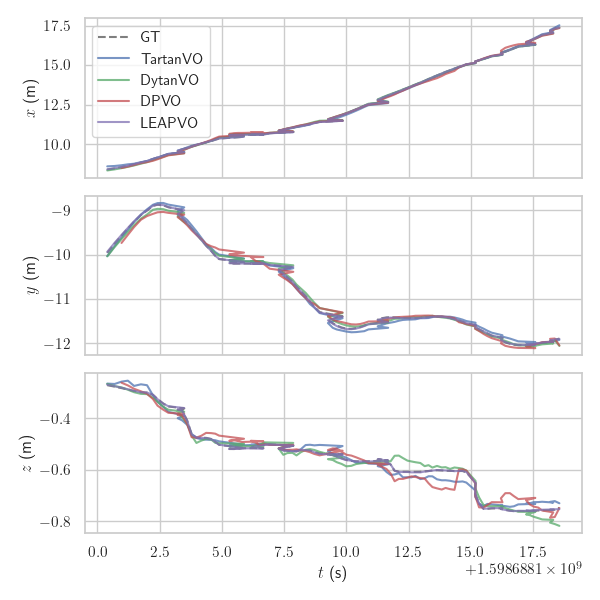}
        \setlength{\abovecaptionskip}{-0.45cm}
         \caption{RoadCrossing 05}
     \end{subfigure}
     \begin{subfigure}[b]{0.245\textwidth}
         \centering
         \includegraphics[width=\textwidth]{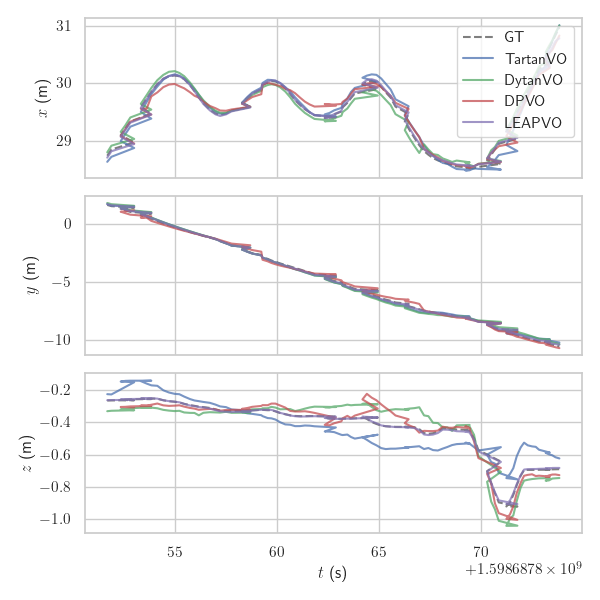}
        \setlength{\abovecaptionskip}{-0.45cm}
         \caption{RoadCrossing 06}
     \end{subfigure}
      \begin{subfigure}[b]{0.245\textwidth}
         \centering
         \includegraphics[width=\textwidth]{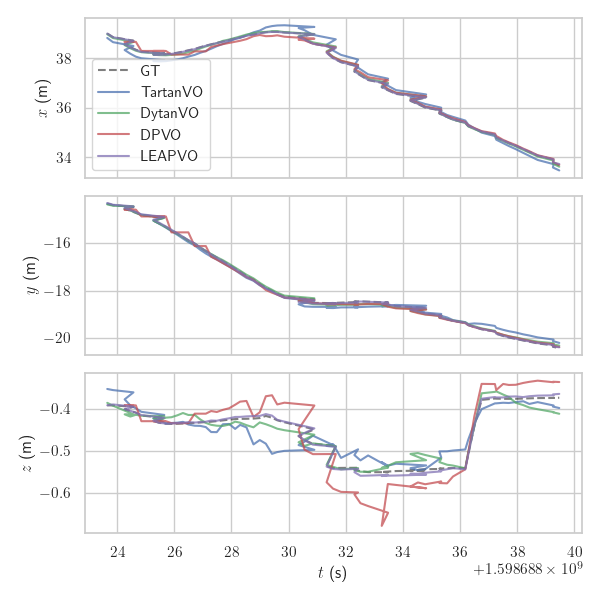}
        \setlength{\abovecaptionskip}{-0.45cm}
         \caption{RoadCrossing 07}
     \end{subfigure}
     
    \caption{
     \textbf{Qualitative results of camera trajectory estimation on TartanAir-Shibuya~\cite{qiu2022airdos}.
    } The visualizations show that our method provides more robust and accurate camera pose trajectories, especially in hard cases (RoadCrossing 06 and RoadCrossing 07). 
    }
    \label{fig:vo_shibuya}
    \vspace{-6pt}
\end{figure*}
\vspace{-5pt}
\subsection{Implementation details}
%
We employ CoTracker~\cite{karaev2023cotracker}, a recent PIPs variant, as our base TAP model. We train our model on the TAP-Vid-Kubric~\cite{doersch2022tap} training set, following~\cite{karaev2023cotracker}. 
We load the pre-trained CoTracker weights for the image feature extractor and train LEAP for 100,000 steps using 4 NVIDIA A100 GPUs in parallel.
During training, we use $K=4$ steps for iterative refinements, $N=256$ for the number of queries, $S=8$ for the model window, and $S_{LP}=12$ for the tracking window.
%
%
For inter-track anchor-based attention, the number of anchors is 64. 
The loss weights are set to $w_1=1.0, w_2=0.5, w_3=0.5$.
In LEAP-VO, we perform 4 steps for each bundle adjustment with the Huber Loss for distance metric. The bundle adjustment window size $S_{BA}$ is set to 15.  
Track filtering parameters are set at $\gamma_v=0.9, \gamma_{d}=0.9, \gamma_u = 0.8, \gamma_{track}=3$. 
%
%
Please refer to our supplementary materials for more details.


\subsection{Visual Odometry Results}
We compare our methods against the state-of-the-art SLAM and VO approaches, including ORB-SLAM2~\cite{murorbslam2}, DynaSLAM~\cite{bescos2018dynaslam}, DROID-SLAM~\cite{teed2021droid}, TartanVO~\cite{wang2021tartanvo}, DytanVO~\cite{shen2023dytanvo}, and DPVO~\cite{teed2022deep}. Evaluations are conducted on the static Replica dataset~\cite{straub2019replica} as well as on the dynamic datasets MPI-Sintel~\cite{Butler2012sintel} and TartanAir-Shibuya~\cite{qiu2022airdos}.\\

\noindent
\textbf{Replica}.
We compare our method with other baselines on the Replica dataset, which contains eight static scenes without any moving objects. Challenges arise in two test sequences that contain prolonged occlusions, leading to the potential camera tracking loss. 
As illustrated in \Cref{tab:vo_replica}, our approach outperforms other VO methods on all three metrics, showing effectiveness under static environments and the capability to handle occlusions.  Further insights are available in \Cref{sec:exp_track_filter}. 
For classical methods, ORB-SLAM2~\cite{murorbslam2} and DynaSLAM~\cite{bescos2018dynaslam}, we observe tracking loss and resuming via re-localization, which reduces long-term drift and shows high accuracy for camera pose tracking. \\[-6pt]

\noindent
\textbf{MPI-Sintel}.
We also evaluate our VO camera trajectory accuracy on 14 dynamic scenes from the MPI Sintel dataset, following the setting of~\cite{zhao2022particlesfm}. As shown in \Cref{tab:vo_sintel}, our method substantially surpasses other VO baselines in dynamic scenes, thanks to our temporal uncertainty estimation and dynamic track detection. Traditional feature-based methods like ORB-SLAM2~\cite{murorbslam2} and DynaSLAM~\cite{bescos2018dynaslam} sometimes fail to track camera movements in highly dynamic sequences, primarily due to outliers from dynamic object correspondences and segmentation-based only dynamic track estimation. In contrast, our approach, which considers visual, temporal, and inter-track information, demonstrates notable improvements in dynamic scene tracking accuracy.\\[-6pt]

\noindent
\textbf{TartanAir-Shibuya}. Following DytanVO~\cite{shen2023dytanvo}, we further evaluate the VO performance using the dynamic TartanAir-Shibuya dataset. We assess the Average Trajectory Error (ATE) across seven diverse scenes, ranging from simple to challenging. 
As depicted in \Cref{tab:vo_shibuya}, our method consistently outperforms all other SLAM and VO baselines in the majority of these scenes.
Moreover, trajectory visualizations in \Cref{fig:vo_shibuya} also highlight the robust camera tracking capabilities of our method in dynamic environments.

\subsection{Dynamic Track Estimation Results}
\begin{figure}[]
 \setlength{\abovecaptionskip}{0.1cm}
 \setlength{\belowcaptionskip}{-0.3cm}
     \centering
      \begin{subfigure}[b]{0.47\textwidth}
         \centering
         \includegraphics[width=\textwidth]{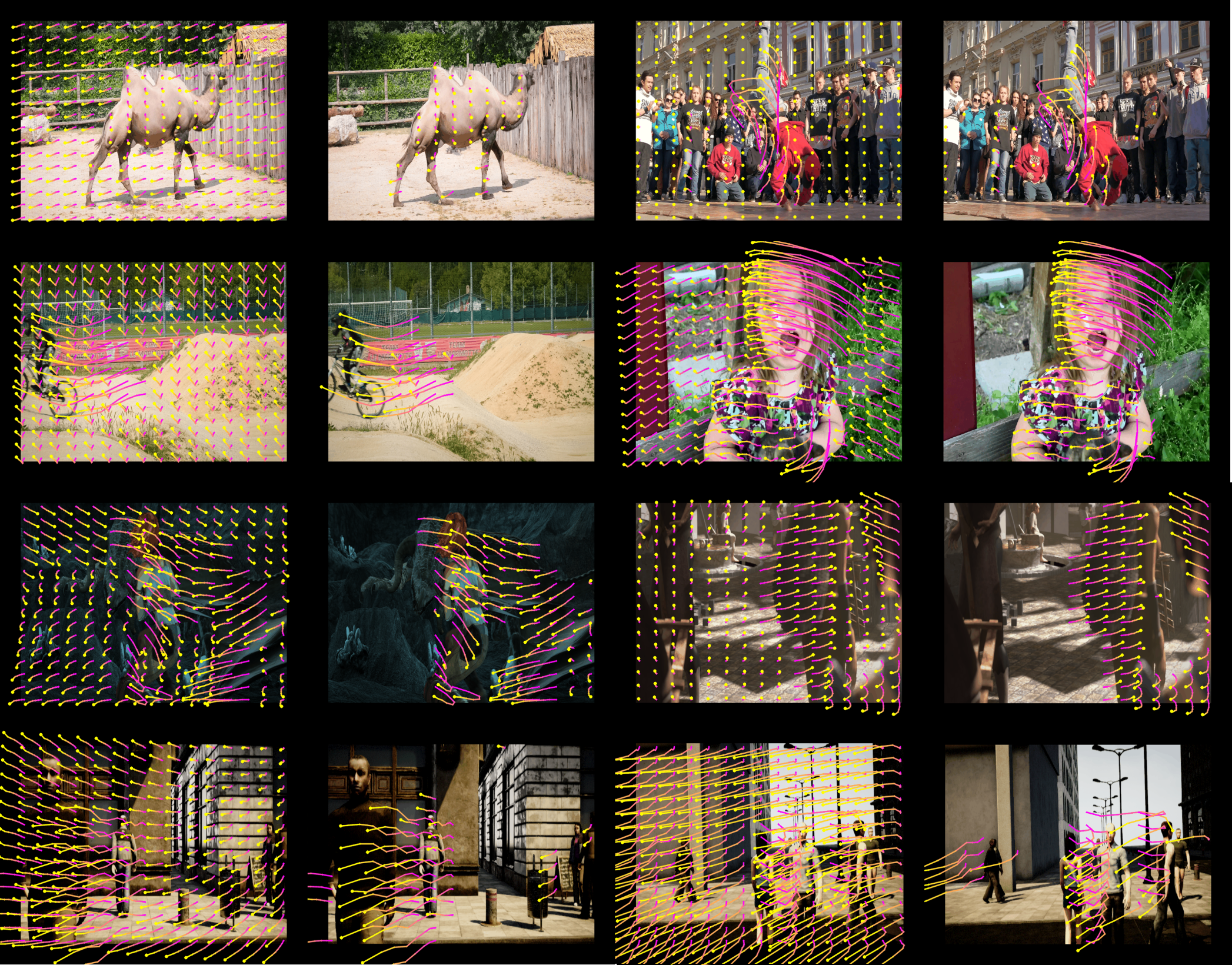}
     \end{subfigure}

    \caption{
    \textbf{Visualization of dynamic track estimation on DAVIS~\cite{KhoRohrSch_davis}, MPI-Sintel~\cite{Butler2012sintel}, and TartanAir-Shibuya~\cite{qiu2022airdos}.} Odd columns: all point trajectories. Even columns:  estimated dynamic point trajectories.
    }
    \label{fig:vis_motion_davis}
\end{figure}

In our LEAP formulation, we have introduced the anchor-based dynamic track estimation module to distinguish between static and dynamic point trajectories, which provides valuable information for handling dynamic scenes in visual odometry. Our method leverages both point feature information and inter-track relations to predict the dynamic track label for each trajectory. To demonstrate the performance of the proposed dynamic track estimation module, we take the LEAP model and evaluate it on novel image sequences from diverse datasets. 
For a comprehensive demonstration, we uniformly sample $16\times 16$ queries among the images and track them for $T=8$ frames. The results, shown in \Cref{fig:vis_motion_davis}, illustrate that LEAP can effectively handle various scenarios, including single-object motion, complex and rapid motion, and multi-object motion with occlusion. 

\subsection{Ablation Study}

\noindent
\textbf{Track Filtering}.
\label{sec:exp_track_filter}
To highlight the efficacy of our track filtering module in LEAP-VO, we conduct an ablation study on different masks used in track filtering. We study three types of filtering criteria: visibility, dynamic track, and confidence filtering. 
\Cref{tab:ablation_track_filter} shows the impact of applying various filtering types on the dynamic MPI Sintel benchmark.
Without dynamic filtering or confidence filtering, the ATE increases by 46\% and 130\%, respectively. Combining all three types of filtering methods altogether, we achieve the best results in all three metrics.\\

\vspace{-0.4cm}
\begin{table}[ht]
\centering
\footnotesize
 \setlength{\abovecaptionskip}{0.1cm}
 \setlength{\belowcaptionskip}{-0.2cm}
\setlength\tabcolsep{6pt}
\vspace{-4pt}
\begin{tabular}{@{}ccccccc@{}}
\toprule
\multirow{2}{*}{Method}  & \multicolumn{3}{c}{Track Filtering} & \multicolumn{3}{c}{MPI Sintel}                   \\ \cmidrule(l){2-7} 
                         & Vis.       & Dyn.      & Conf.      & ATE       & RPE trans & RPE rot   \\ \midrule
\multirow{4}{*}{LEAP-VO} 
 & \checkmark          & -         & -          &   0.118       &    0.078      &   1.340        \\
                         & \checkmark          & \checkmark         & -          & 0.085        &  0.062       & 1.281      \\
                         & \checkmark          & -         & \checkmark          &  0.054         &  0.058       &  1.274      \\
                         & \checkmark          & \checkmark         & \checkmark          &   \textbf{0.037} & \textbf{0.055} & \textbf{1.263}  \\ \bottomrule
\end{tabular}
\caption{\textbf{Effect of the proposed track filtering method on MPI-Sintel~\cite{Butler2012sintel}.} Tracking using all criteria yields the best results. }
\label{tab:ablation_track_filter}
\vspace{-6pt}
\end{table}

\noindent
\textbf{Uncertainty Estimation}.
\begin{figure}[]
 \setlength{\abovecaptionskip}{0.1cm}
     \centering
   \begin{subfigure}[]{0.47\textwidth}
         \centering
         \includegraphics[width=\textwidth]{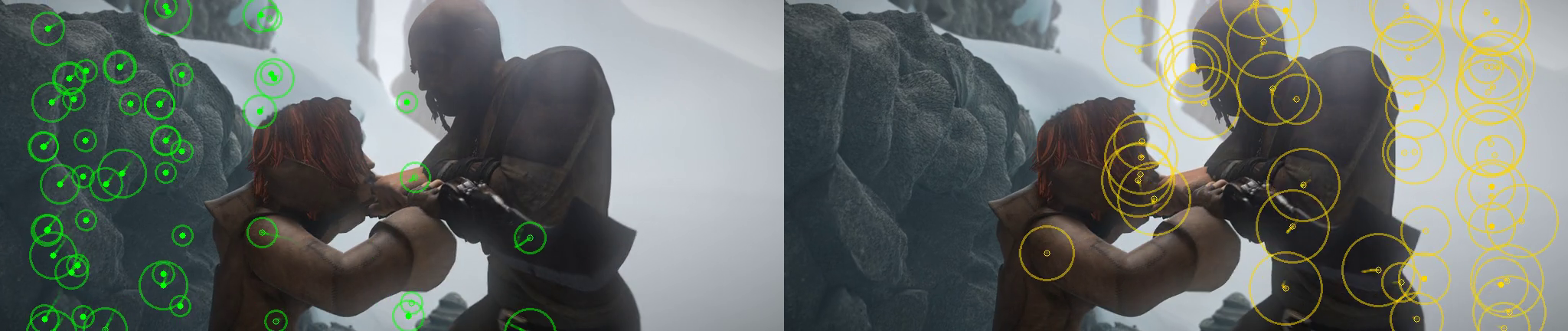}
     \end{subfigure}
      \begin{subfigure}[]{0.47\textwidth}
         \centering
         \includegraphics[width=\textwidth]{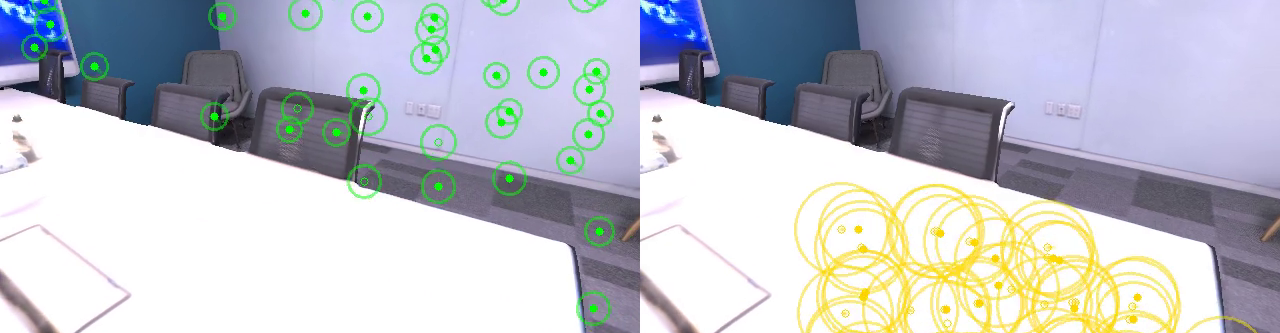}
     \end{subfigure}
     
    \caption{
    \textbf{Visualization of point-wise uncertainty measurements.} Keypoints with the lowest 20\% (left) and highest 20\% (right) uncertainty are shown in green and yellow, respectively.
    }
    \label{fig:uncertainty_vis}
    \vspace{-0.5cm}
\end{figure}
As illustrated in \Cref{fig:uncertainty_vis}, we demonstrate the LEAP's capability to provide meaningful per-point uncertainty.
It models the uncertainty as distribution parameters for both the entire track and individual points. Two key observations can be found from \Cref{fig:uncertainty_vis}. Firstly, our model reliably identifies high-uncertainty points, typically located in low-texture areas or in regions with repetitive patterns. 
Secondly, although not explicitly supervised, the model tends to assign higher uncertainty to dynamic objects, which are inherently more challenging to track.
These findings highlight the efficacy of LEAP's probabilistic formulation in handling uncertainty.\\

\vspace{-4pt}

\noindent
\textbf{Keypoint Extraction}.
\Cref{tab:ablation_anchor} showcases the effectiveness of various keypoint extraction strategies for VO performance. 
Despite its distinctive features, SIFT extraction provides inferior performance due to its unpredictable keypoint distribution.
In contrast, the random sampling tends to generate well-separated keypoints, leading to a performance similar to SIFT's.
Our distributed image gradient-based sampling shows the best performance, underscoring the significance of feature distribution and distinctiveness.

\vspace{-0.2cm}
\begin{table}[h]
\centering
\footnotesize
 \setlength{\abovecaptionskip}{0.1cm}
 \setlength{\belowcaptionskip}{-0.2cm}
\setlength\tabcolsep{5pt}
\vspace{-4pt}
\begin{tabular}{@{}ccccc@{}}
\toprule
Policy:      & Uniform & Random & SIFT~\cite{lowe1999sift}   & Ours   \\ \midrule
ATE (m): & 0.1511  & 0.0366 & 0.0355 & \textbf{0.0290} \\ \bottomrule
\end{tabular}
\caption{\textbf{Comparison of keypoint extraction policies ($N_a=64$) on TartanAir-Shibuya~\cite{qiu2022airdos}.} Our distributed image-gradient sampling method achieves better results compared to other policies.}
\label{tab:ablation_anchor}
\vspace{-4pt}
\end{table}


\noindent
\textbf{Baseline comparison}. 
We evaluate the pretrained CoTracker~\cite{karaev2023cotracker} for the end VO performance. 
\Cref{tab:ablation_cotracker} shows that LEAP front-end outperforms CoTracker on all datasets, demonstrating the effectiveness of the proposed dynamic track estimation and temporal probability modeling.
\begin{table}[ht]
\centering
\footnotesize
\setlength{\abovecaptionskip}{0.1cm}
\setlength{\belowcaptionskip}{-0.2cm}
\setlength\tabcolsep{1pt}
\vspace{-4pt}
\begin{tabular}{@{}lccccccc@{}}
\toprule
\multirow{2}{*}{Front-end} & \multicolumn{3}{c}{Replica}                      & \multicolumn{3}{c}{MPI Sintel}                   & Shibuya        \\ \cmidrule(l){2-8} 
                           & ATE            & RPE trans      & RPE rot        & ATE            & RPE trans      & RPE rot        & ATE            \\ \midrule
CoTracker                 & 0.301          & 0.031          & 2.043          & 0.126          & 0.071          & 1.366          & 0.160          \\
LEAP                       & \textbf{0.204} & \textbf{0.030} & \textbf{1.992} & \textbf{0.037} & \textbf{0.055} & \textbf{1.263} & \textbf{0.029} \\ \bottomrule
\end{tabular}
\vspace{-2pt}
\caption{{\textbf{Baseline comparison with CoTracker front-end~\cite{karaev2023cotracker}.}} Our LEAP front-end achieves better results on all datasets.}
\label{tab:ablation_cotracker}
\vspace{-4pt}
\end{table}
\vspace{-0.2cm}

\section{Conclusion}
\label{sec:conclusion}
We introduce Long-term Effective Any Point Tracking (LEAP), designed to tackle the limitations of previous point-tracking methods. By leveraging visual, inter-track, and temporal cues within an anchor-based dynamic track estimation module, LEAP captures global motion patterns and effectively discriminates between static and dynamic elements. Moreover, our approach incorporates a temporal probabilistic formulation into the point-tracking pipeline, emulating recursive Bayesian filtering through a learning-based refinement module. This enables our model to accurately assess the reliability of point trajectory estimates.
Using LEAP as the front-end, we then develop the LEAP-VO system, featuring novel integration of visual information, global motion patterns, and track distribution for handling dynamic scenes. 
Additionally, our takes for LEAP-VO can be integrated with other TAP-based front-ends, offering the potential to markedly improve camera-tracking accuracy and robustness.

{
    \small
    \bibliographystyle{ieeenat_fullname}
    \bibliography{main}

@String(CVPR= {IEEE Conf. Comput. Vis. Pattern Recog.})

@String(ICCV= {Int. Conf. Comput. Vis.})

@String(ECCV= {Eur. Conf. Comput. Vis.})

@String(ACCV  = {ACCV})

@String(ICLR = {Int. Conf. Learn. Represent.})

@String(CVPR  = {CVPR})

@String(ICCV  = {ICCV})

@String(ECCV  = {ECCV})

@String(ICLR  = {ICLR})

@inproceedings{harley2022particle,
  title={Particle video revisited: Tracking through occlusions using point trajectories},
  author={Harley, Adam W and Fang, Zhaoyuan and Fragkiadaki, Katerina},
  booktitle={ECCV},
  pages={59--75},
  year={2022},
  organization={Springer}
}

@article{karaev2023cotracker,
  title={Cotracker: It is better to track together},
  author={Karaev, Nikita and Rocco, Ignacio and Graham, Benjamin and Neverova, Natalia and Vedaldi, Andrea and Rupprecht, Christian},
  journal={arXiv preprint arXiv:2307.07635},
  year={2023}
}

@article{doersch2023tapir,
  title={TAPIR: Tracking Any Point with per-frame Initialization and temporal Refinement},
  author={Doersch, Carl and Yang, Yi and Vecerik, Mel and Gokay, Dilara and Gupta, Ankush and Aytar, Yusuf and Carreira, Joao and Zisserman, Andrew},
  journal={ICCV},
  year={2023}
}

@inproceedings{zhang2022structure,
  title={Structure and motion from casual videos},
  author={Zhang, Zhoutong and Cole, Forrester and Li, Zhengqi and Rubinstein, Michael and Snavely, Noah and Freeman, William T},
  booktitle={ECCV},
  pages={20--37},
  year={2022},
  organization={Springer}
}

@inproceedings{zhao2022particlesfm,
  title={Particlesfm: Exploiting dense point trajectories for localizing moving cameras in the wild},
  author={Zhao, Wang and Liu, Shaohui and Guo, Hengkai and Wang, Wenping and Liu, Yong-Jin},
  booktitle={ECCV},
  pages={523--542},
  year={2022},
  organization={Springer}
}

@inproceedings{kopf2021robust,
  title={Robust consistent video depth estimation},
  author={Kopf, Johannes and Rong, Xuejian and Huang, Jia-Bin},
  booktitle={CVPR},
  pages={1611--1621},
  year={2021}
}

@inproceedings{lowe1999sift,
  title={Object recognition from local scale-invariant features},
  author={Lowe, David G},
  booktitle={ICCV},
  volume={2},
  pages={1150--1157},
  year={1999},
  organization={Ieee}
}

@article{teed2022deep,
  title={Deep patch visual odometry},
  author={Teed, Zachary and Lipson, Lahav and Deng, Jia},
  journal={NeurIPS},
  volume={36},
  year={2024}
}

@inproceedings{triggs1999bundle,
  title={Bundle adjustment—a modern synthesis},
  author={Triggs, Bill and McLauchlan, Philip F and Hartley, Richard I and Fitzgibbon, Andrew W},
  booktitle={International workshop on vision algorithms},
  pages={298--372},
  year={1999},
  organization={Springer}
}

@inproceedings{teed2020raft,
  title={Raft: Recurrent all-pairs field transforms for optical flow},
  author={Teed, Zachary and Deng, Jia},
  booktitle={ECCV},
  pages={402--419},
  year={2020},
  organization={Springer}
}

@article{doersch2022tap,
  title={Tap-vid: A benchmark for tracking any point in a video},
  author={Doersch, Carl and Gupta, Ankush and Markeeva, Larisa and Recasens, Adri{\`a} and Smaira, Lucas and Aytar, Yusuf and Carreira, Jo{\~a}o and Zisserman, Andrew and Yang, Yi},
  journal={NeurIPS},
  volume={35},
  pages={13610--13626},
  year={2022}
}

@article{bian2023contexttap,
  title={Context-TAP: Tracking Any Point Demands Spatial Context Features},
  author={Bian, Weikang and Huang, Zhaoyang and Shi, Xiaoyu and Dong, Yitong and Li, Yijin and Li, Hongsheng},
  journal={arXiv preprint arXiv:2306.02000},
  year={2023}
}

@article{wang2023omnimotion,
  title={Tracking Everything Everywhere All at Once},
  author={Wang, Qianqian and Chang, Yen-Yu and Cai, Ruojin and Li, Zhengqi and Hariharan, Bharath and Holynski, Aleksander and Snavely, Noah},
  journal={arXiv preprint arXiv:2306.05422},
  year={2023}
}

@article{jabri2020space,
  title={Space-time correspondence as a contrastive random walk},
  author={Jabri, Allan and Owens, Andrew and Efros, Alexei},
  journal={NeurIPS},
  volume={33},
  pages={19545--19560},
  year={2020}
}

@inproceedings{caron2021emerging,
  title={Emerging properties in self-supervised vision transformers},
  author={Caron, Mathilde and Touvron, Hugo and Misra, Ishan and J{\'e}gou, Herv{\'e} and Mairal, Julien and Bojanowski, Piotr and Joulin, Armand},
  booktitle={CVPR},
  pages={9650--9660},
  year={2021}
}

@article{davison2007monoslam,
  title={MonoSLAM: Real-time single camera SLAM},
  author={Davison, Andrew J and Reid, Ian D and Molton, Nicholas D and Stasse, Olivier},
  journal={IEEE TPAMI},
  volume={29},
  number={6},
  pages={1052--1067},
  year={2007},
  publisher={IEEE}
}

@article{mur2015orb,
  title={ORB-SLAM: a versatile and accurate monocular SLAM system},
  author={Mur-Artal, Raul and Montiel, Jose Maria Martinez and Tardos, Juan D},
  journal={IEEE Transactions on Robotics},
  volume={31},
  number={5},
  pages={1147--1163},
  year={2015},
  publisher={IEEE}
}

@inproceedings{wang2017deepvo,
  title={Deepvo: Towards end-to-end visual odometry with deep recurrent convolutional neural networks},
  author={Wang, Sen and Clark, Ronald and Wen, Hongkai and Trigoni, Niki},
  booktitle={ICRA},
  pages={2043--2050},
  year={2017},
  organization={IEEE}
}

@inproceedings{zhou2017unsupervised,
  title={Unsupervised learning of depth and ego-motion from video},
  author={Zhou, Tinghui and Brown, Matthew and Snavely, Noah and Lowe, David G},
  booktitle={CVPR},
  pages={1851--1858},
  year={2017}
}

@inproceedings{wang2021tartanvo,
  title={Tartanvo: A generalizable learning-based vo},
  author={Wang, Wenshan and Hu, Yaoyu and Scherer, Sebastian},
  booktitle={CoRL},
  pages={1761--1772},
  year={2021},
  organization={PMLR}
}

@inproceedings{engel2014lsd,
  title={LSD-SLAM: Large-scale direct monocular SLAM},
  author={Engel, Jakob and Sch{\"o}ps, Thomas and Cremers, Daniel},
  booktitle={ECCV},
  pages={834--849},
  year={2014},
  organization={Springer}
}

@article{engel2017direct,
  title={Direct sparse odometry},
  author={Engel, Jakob and Koltun, Vladlen and Cremers, Daniel},
  journal={IEEE TPAMI},
  volume={40},
  number={3},
  pages={611--625},
  year={2017},
  publisher={IEEE}
}

@article{teed2021droid,
  title={Droid-slam: Deep visual slam for monocular, stereo, and rgb-d cameras},
  author={Teed, Zachary and Deng, Jia},
  journal={NeurIPS},
  volume={34},
  pages={16558--16569},
  year={2021}
}

@inproceedings{wang2017stereo,
  title={Stereo DSO: Large-scale direct sparse visual odometry with stereo cameras},
  author={Wang, Rui and Schworer, Martin and Cremers, Daniel},
  booktitle={ICCV},
  pages={3903--3911},
  year={2017}
}

@article{straub2019replica,
  title={The Replica dataset: A digital replica of indoor spaces},
  author={Straub, Julian and Whelan, Thomas and Ma, Lingni and Chen, Yufan and Wijmans, Erik and Green, Simon and Engel, Jakob J and Mur-Artal, Raul and Ren, Carl and Verma, Shobhit and others},
  journal={arXiv preprint arXiv:1906.05797},
  year={2019}
}

@inproceedings{zhi2021place,
  title={In-place scene labelling and understanding with implicit scene representation},
  author={Zhi, Shuaifeng and Laidlow, Tristan and Leutenegger, Stefan and Davison, Andrew J},
  booktitle={CVPR},
  pages={15838--15847},
  year={2021}
}

@inproceedings{KhoRohrSch_davis,
                title = {Video Object Segmentation with Language Referring Expressions},
                author = {Khoreva, Anna and Rohrbach, Anna and Schiele, Bernt},
                year = {2018},
                booktitle = {ACCV}
                }

@inproceedings{Butler2012sintel,
title = {A naturalistic open source movie for optical flow evaluation},
author = {Butler, D. J. and Wulff, J. and Stanley, G. B. and Black, M. J.},
booktitle = {ECCV},
editor = {{A. Fitzgibbon et al. (Eds.)}},
publisher = {Springer-Verlag},
series = {Part IV, LNCS 7577},
month = oct,
pages = {611--625},
year = {2012}
}

@article{fischler1981ransac,
  title={Random sample consensus: a paradigm for model fitting with applications to image analysis and automated cartography},
  author={Fischler, Martin A and Bolles, Robert C},
  journal={Communications of the ACM},
  volume={24},
  number={6},
  pages={381--395},
  year={1981},
  publisher={ACM New York, NY, USA}
}

@article{murorbslam2,
  title={{ORB-SLAM2}: an Open-Source {SLAM} System for Monocular, Stereo and {RGB-D} Cameras},
  author={Mur-Artal, Ra\'ul and Tard\'os, Juan D.},
  journal={IEEE Transactions on Robotics},
  volume={33},
  number={5},
  pages={1255--1262},
  doi = {10.1109/TRO.2017.2705103},
  year={2017}
 }

@inproceedings{shen2023dytanvo,
  title={DytanVO: Joint refinement of visual odometry and motion segmentation in dynamic environments},
  author={Shen, Shihao and Cai, Yilin and Wang, Wenshan and Scherer, Sebastian},
  booktitle={ICRA},
  pages={4048--4055},
  year={2023},
  organization={IEEE}
}

@article{bescos2018dynaslam,
  title={DynaSLAM: Tracking, mapping, and inpainting in dynamic scenes},
  author={Bescos, Berta and F{\'a}cil, Jos{\'e} M and Civera, Javier and Neira, Jos{\'e}},
  journal={IEEE Robotics and Automation Letters},
  volume={3},
  number={4},
  pages={4076--4083},
  year={2018},
  publisher={IEEE}
}

@inproceedings{qiu2022airdos,
  title={AirDOS: Dynamic SLAM benefits from articulated objects},
  author={Qiu, Yuheng and Wang, Chen and Wang, Wenshan and Henein, Mina and Scherer, Sebastian},
  booktitle={ICRA},
  pages={8047--8053},
  year={2022},
  organization={IEEE}
}

@article{chen2023uncertainty,
  title={Uncertainty-Driven Dense Two-View Structure From Motion},
  author={Chen, Weirong and Kumar, Suryansh and Yu, Fisher},
  journal={IEEE Robotics and Automation Letters},
  volume={8},
  number={3},
  pages={1763--1770},
  year={2023},
  publisher={IEEE}
}

@inproceedings{yang2020d3vo,
  title={D3vo: Deep depth, deep pose and deep uncertainty for monocular visual odometry},
  author={Yang, Nan and Stumberg, Lukas von and Wang, Rui and Cremers, Daniel},
  booktitle={CVPR},
  pages={1281--1292},
  year={2020}
}

@inproceedings{yin2018geonet,
  title={Geonet: Unsupervised learning of dense depth, optical flow and camera pose},
  author={Yin, Zhichao and Shi, Jianping},
  booktitle={CVPR},
  pages={1983--1992},
  year={2018}
}

@inproceedings{ranjan2019competitive,
  title={Competitive collaboration: Joint unsupervised learning of depth, camera motion, optical flow and motion segmentation},
  author={Ranjan, Anurag and Jampani, Varun and Balles, Lukas and Kim, Kihwan and Sun, Deqing and Wulff, Jonas and Black, Michael J},
  booktitle={CVPR},
  pages={12240--12249},
  year={2019}
}

@inproceedings{bloesch2018codeslam,
  title={Codeslam—learning a compact, optimisable representation for dense visual slam},
  author={Bloesch, Michael and Czarnowski, Jan and Clark, Ronald and Leutenegger, Stefan and Davison, Andrew J},
  booktitle={CVPR},
  pages={2560--2568},
  year={2018}
}

@inproceedings{chen2019self,
  title={Self-supervised learning with geometric constraints in monocular video: Connecting flow, depth, and camera},
  author={Chen, Yuhua and Schmid, Cordelia and Sminchisescu, Cristian},
  booktitle={ICCV},
  pages={7063--7072},
  year={2019}
}

@inproceedings{mossel2016streaming,
  title={Streaming and exploration of dynamically changing dense 3d reconstructions in immersive virtual reality},
  author={Mossel, Annette and Kroeter, Manuel},
  booktitle={2016 IEEE International Symposium on Mixed and Augmented Reality (ISMAR-Adjunct)},
  pages={43--48},
  year={2016},
  organization={IEEE}
}

@article{yousif2015overview,
  title={An overview to visual odometry and visual SLAM: Applications to mobile robotics},
  author={Yousif, Khalid and Bab-Hadiashar, Alireza and Hoseinnezhad, Reza},
  journal={Intelligent Industrial Systems},
  volume={1},
  number={4},
  pages={289--311},
  year={2015},
  publisher={Springer}
}

@article{geiger2013vision,
  title={Vision meets robotics: The kitti dataset},
  author={Geiger, Andreas and Lenz, Philip and Stiller, Christoph and Urtasun, Raquel},
  journal={The International Journal of Robotics Research},
  volume={32},
  number={11},
  pages={1231--1237},
  year={2013},
  publisher={Sage Publications Sage UK: London, England}
}

@inproceedings{zheng2023pointodyssey,
  title={Pointodyssey: A large-scale synthetic dataset for long-term point tracking},
  author={Zheng, Yang and Harley, Adam W and Shen, Bokui and Wetzstein, Gordon and Guibas, Leonidas J},
  booktitle={ICCV},
  pages={19855--19865},
  year={2023}
}

@article{doersch2024bootstap,
  title={BootsTAP: Bootstrapped Training for Tracking-Any-Point},
  author={Doersch, Carl and Yang, Yi and Gokay, Dilara and Luc, Pauline and Koppula, Skanda and Gupta, Ankush and Heyward, Joseph and Goroshin, Ross and Carreira, Jo{\~a}o and Zisserman, Andrew},
  journal={arXiv preprint arXiv:2402.00847},
  year={2024}
}

@article{tumanyan2024dino,
  title={DINO-Tracker: Taming DINO for Self-Supervised Point Tracking in a Single Video},
  author={Tumanyan, Narek and Singer, Assaf and Bagon, Shai and Dekel, Tali},
  journal={arXiv preprint arXiv:2403.14548},
  year={2024}
}

@inproceedings{ye2023pvo,
  title={PVO: Panoptic visual odometry},
  author={Ye, Weicai and Lan, Xinyue and Chen, Shuo and Ming, Yuhang and Yu, Xingyuan and Bao, Hujun and Cui, Zhaopeng and Zhang, Guofeng},
  booktitle={CVPR},
  pages={9579--9589},
  year={2023}
}

@article{wu2022d,
  title={D\^{} 2nerf: Self-supervised decoupling of dynamic and static objects from a monocular video},
  author={Wu, Tianhao and Zhong, Fangcheng and Tagliasacchi, Andrea and Cole, Forrester and Oztireli, Cengiz},
  journal={NeurIPS},
  volume={35},
  pages={32653--32666},
  year={2022}
}

@inproceedings{tschernezki2021neuraldiff,
  title={NeuralDiff: Segmenting 3D objects that move in egocentric videos},
  author={Tschernezki, Vadim and Larlus, Diane and Vedaldi, Andrea},
  booktitle={3DV},
  pages={910--919},
  year={2021},
  organization={IEEE}
}

@article{yang2023emernerf,
  title={Emernerf: Emergent spatial-temporal scene decomposition via self-supervision},
  author={Yang, Jiawei and Ivanovic, Boris and Litany, Or and Weng, Xinshuo and Kim, Seung Wook and Li, Boyi and Che, Tong and Xu, Danfei and Fidler, Sanja and Pavone, Marco and others},
  journal={arXiv preprint arXiv:2311.02077},
  year={2023}
}

@inproceedings{teed2019deepv2d,
  title={DeepV2D: Video to Depth with Differentiable Structure from Motion},
  author={Teed, Zachary and Deng, Jia},
  booktitle={ICLR},
  year={2019}
}

@inproceedings{li2020self,
  title={Self-supervised deep visual odometry with online adaptation},
  author={Li, Shunkai and Wang, Xin and Cao, Yingdian and Xue, Fei and Yan, Zike and Zha, Hongbin},
  booktitle={CVPR},
  pages={6339--6348},
  year={2020}
}
}



\end{document}